\documentclass[sigconf,nonacm]{acmart}

\AtBeginDocument{%
 }

\setcopyright{none}
\usepackage{booktabs}
\usepackage{amsmath}
\usepackage{algorithm}
\usepackage{algpseudocode}
\usepackage{graphicx}

\graphicspath{{figs/}{figures/}{pictures/}{images/}{./}}

\author{Bowen Yu}
\affiliation{%
  \institution{Wenzhou-Kean University}
  \city{Wenzhou}
  \country{China}
}
\email{yubow@kean.edu}

\author{Mingyu Huang}
\affiliation{%
  \institution{Wenzhou-Kean University}
  \city{Wenzhou}
  \country{China}
}
\email{huanming@kean.edu}

\author{Yishen Liu}
\affiliation{%
  \institution{Wenzhou-Kean University}
  \city{Wenzhou}
  \country{China}
}
\email{liuyis@kean.edu}

\author{Yue Zhao}
\affiliation{%
  \institution{Wenzhou-Kean University}
  \city{Wenzhou}
  \country{China}
}
\email{yuezhao@kean.edu}

\title{When Speech Meets Lips: Interpretable Audio-Visual Synchronization for L2 Pronunciation Assessment}

\begin{document}

\begin{abstract}
Automatic Pronunciation Assessment (APA) systems have achieved remarkable performance via transformer-based models and self-supervised speech representations. However, most methods rely solely on acoustic signals, overlooking cross-modal temporal synchronization between speech and articulatory movements, thus providing insufficient diagnostic feedback on timing mismatches critical for L2 pronunciation training. To address this, we propose an interpretable audio--visual synchronization framework explicitly modeling speech--lip temporal alignment, including feature encoding, cross-attention fusion, lag estimation, stability quantification, and visualization. We introduce frame-level lag trajectories and a Lag Stability Index (LSI) to quantify synchronization robustness. In addition, we interviewed 30 participants (10 instructors, 20 diverse L1 students) to validate the framework's effectiveness. Our framework transforms implicit alignment into interpretable representations, bridging automatic scoring and actionable Computer-Aided Pronunciation Training (CAPT) feedback. Datasets of this paper and all supplemental materials are available at \url{https://www.robots.ox.ac.uk/~vgg/data/lip_reading/}.
\end{abstract}


\begin{teaserfigure}
 \centering
 \includegraphics[width=\textwidth]{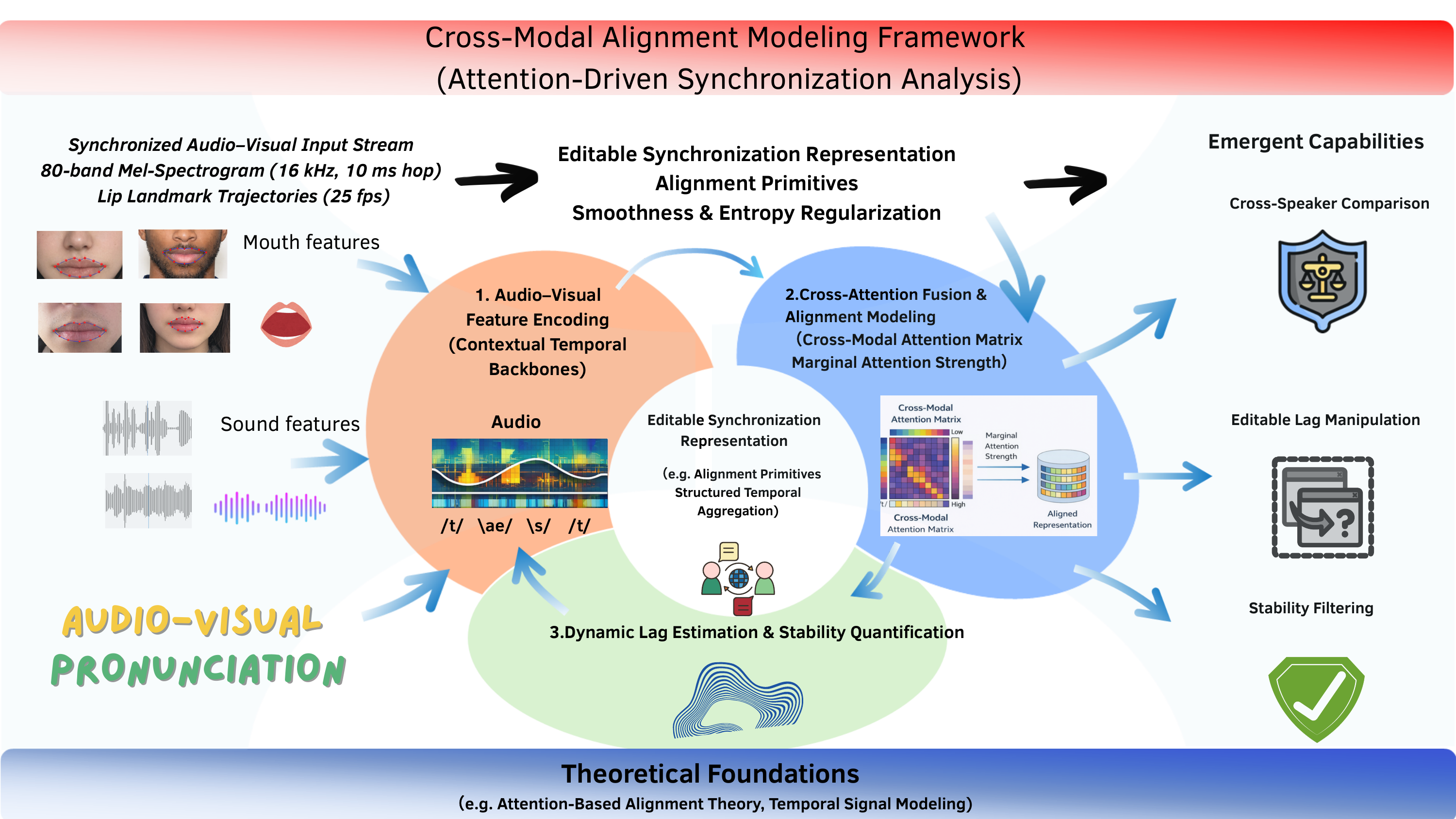}
 \caption{Overview of the proposed audio–visual synchronization analysis framework. The system models cross-modal temporal alignment between speech and lip movements through feature encoding, cross-attention fusion, lag estimation, and stability quantification, enabling interpretable pronunciation feedback.}
 \Description{Overview of the proposed audio–visual synchronization analysis framework.}
 \label{fig:teaser}
\end{teaserfigure}

\maketitle

\section{Introduction}

Pronunciation is a form of hearing, so some aspects of observation should be included in terms of the presentation of sounds. During daily conversations, although listeners focus more on the speaker's body language at their mouths during speeches, in a noisy environment with low volume under noisy conditions or poor voice status among many people, they will pay greater attention to eye movement information conveyed by visual stimuli on faces for identification purposes. When sound and lip movement are inconsistent - for instance, due to a delay in video calls, it becomes apparent that there is an error; at this time, the speech may seem unclear. Secondly, for students learning a foreign language at present, the time pressure and coordination problems faced by them can also be explained as ``poor pronunciation,'' but attention should be paid to the fact that there is a slight presence of the desired sound~\cite{Venezia2015LexicalReading}.

Automatic Pronunciation Assessment (APA) provides learners with quantified auditory feedback, and these models have gained considerable strength recently in their ability to provide this type of information. Transformer-based scores and end-to-end mispronunciation detection systems have been proven to be more robust than the older alignment-heavy pipelines in their ability to capture rich information and cover longer contexts~\cite{Wu2021TransformerMPD,Yan2024HierarchicalTransformers,Chao2025SelectiveSSM,Gong2022MultiGranularityAPA}. Public resources, for example, SpeechOcean762, can help enhance the reproducibility of evaluations in non native environments~\cite{Zhang2021Speechocean762}. Additionally, most widely used APA systems are based on a single-channel audio input; mainly relying on acoustics to assess the pronunciation quality, and some have employed self-supervised speech representation fine-tuning techniques or decreased label demands~\cite{Kim2022SelfSupervisedAPA,Zahran2023FineTuningSSL,Baevski2020SSLASR}.

Many of the learner problems exist from multiple aspects; primarily, there are phonetic problems, but more often than not, they involve other aspects as well. Coordination issues: The lips have moved before or after the acoustic signal; There is a discrepancy between the timing of articulation and the desired acoustic feature; Or no change in visibility due to anticipated sound has been detected. Since audio-only models cannot take into account such factors, although they may have achieved certain accuracy in a particular case, they cannot provide an assistive reason that is helpful to learners throughout their learning process. Audio-visual model building is appealing yet difficult. Synchronization is not always a fixed deviation of time; speaking speed may show a non-linear trend of fluctuation, and during actual recording, head movements, camera noise, and changes in the observer's position may occur. Work on audio-visual synchronization in the field and active speaker settings shows that there are still problems with alignment when there is no clear rule~\cite{Chen2021AVSyncWild,Wuerkaixi2022AVSyncASD}. On the other hand, modern cross-modal attention models have implicit capabilities to learn alignments~\cite{Liu2025ALIGNVSR,Ahn2024SyncVSR,Diao2023GraphSecurity}. While there may be an outward appearance of alignment in a network, it cannot be detected by either teachers or students.

As the first grade indicator of cross-modal timing in this project. A weakly supervised audio-visual pronunciation evaluation system is proposed, integrating the following five elements: (1) Audio-visual feature encoding; (2) Cross attention fusion; (3) Dynamic lag estimation; (4) Stability quantification; And coordinated visual analysis technology has been introduced for diagnosis. Cross attention is explicitly able to model cross-modal temporal dependencies and produces an attention matrix that precisely exhibits the relationship between acoustic frames and lip movement~\cite{Liu2025ALIGNVSR,Ahn2024SyncVSR}. For ease of understanding, a one-to-one correspondence is established between the mapping. Marginal attention strength and Local attention concentration are also introduced. Furthermore, the structure of Attention has been converted into a Time-evolving Synchronization Trajectory (lags over time). Finally, the Lag Stability Index (LSI) is introduced to summarize the stability of estimation for lags in an utterance; thus, timing robustness can be compared across different speakers or conditions. Finally, Linked views include attention heatmaps, lag trajectory curves, consistency curves of speech, lip geometry dynamic processes, embedding projections, etc., to facilitate interactive pronunciation diagnosis.

Increasing the rate of successful shots can be used as a reference to propose further improvements. As for the above directions of the CAPT system, they also require that the feedback be actionable and user-friendly; therefore, it includes audiovisual correction functions, such as~\cite{Bu2021PTeacher}. That is to say, we associate the quality of pronunciation with an explicit and observable form of audio-visual temporal consistency; Also, we provide such kind of information in a way convenient for judgment and diagnosis under limited human assistance.

\section{Related Work}

\subsection{Automatic pronunciation assessment and mispronunciation detection}

As the neural APA model ages, it is likely to have a more flexible function of learning rich context and improve the accuracy of error identification. Attentive methods can be used to identify segments of incorrect pronunciation in~\cite{Lopez2021AttentionPED}, and a multi-encoder evaluation mechanism based on multiple perspectives of the signal has also appeared~\cite{Lin2021MultiEncoderAPA}. End-to-end mispronunciation detection and diagnosis based on transformer-based technology also reduces dependence on manually designed alignments; However, they can detect longer dependency relationships~\cite{Wu2021TransformerMPD}. Recently, research has been exploring hierarchical Transformers and selective state space style architectures to realize the efficient multi-affordability of pronunciation evaluation~\cite{Yan2024HierarchicalTransformers,Chao2025SelectiveSSM}, and a multidimensional aspect, a multi-granular scoring system that aligns more closely with humans' rating separation of pronunciation dimensions~\cite{Gong2022MultiGranularityAPA,Do2024FeatureMixup,Do2023HierarchicalAPA}. In terms of the scenario of multiple tasks, for example, multi PA~\cite{Chen2024MultiPA}, although there is an obvious impact on scoring pronunciation, at the same time, it can maintain the model's stability when processing different contents.

Regarding the progress of APA, it is also closely related to data and representation learning. SpeechOcean762 provides a large-scale open corpus for research on non-native English pronunciation~\cite{Zhang2021Speechocean762}. Self-supervised speech representations have become increasingly popular as backbone models for audio classification to improve performance without a substantial amount of labeled data~\cite{Kim2022SelfSupervisedAPA}. Recently, some research results have merged the features of Wav2Vec2 for mispronunciation detection and transformation-based Group of Pictures (GOP) modeling to assess intelligibility and pronunciation quality~\cite{Shekar2023Wav2vec2}. A type of fine-tuning of self-supervised models to achieve end-to-end pronunciation evaluation~\cite{Zahran2023FineTuningSSL}. It is a scaling issue, and evidence shows that the self-supervised pretraining method performs well in speech model construction, etc~\cite{Baevski2020SSLASR}. Another path is as follows: TAC-based phoneme-level evaluation~\cite{Cao2024PhonemeCTC}; Hybrid CTC attention design~\cite{Baranwal2022HybridCTC}; Multi-distribution neural model for detection and diagnosis~\cite{Li2017MultidistributionDNN}; Multifunctional or multi-modal extension of multi-character recognition technology to solve the mispronunciation diagnosis problem~\cite{Guo2023Squeezeformer}. Several recent papers have also pointed out the need for interpretability; For instance, multi-task pretraining was introduced to build an interpretable L2 assessment environment and a phonological level Diagnosis Environment~\cite{Chao2025MultiTaskPretraining,Shahin2024PhonologicalLevel}.

\subsection{Audio-Visual Speech Representation Learning}

Multi-modal fusion of audio-visual learning under limited supervision. Self-supervised methods can directly learn audio-visual objects from raw videos~\cite{Afouras2020SelfSupervisedAV}, and they can also learn speech representations via masked multi-modal objectives~\cite{Shi2022MaskedCluster}. AV-data2vec extends the contextualized target prediction in the context of audio-visual data to enhance self-supervision for speech~\cite{Lian2023AVdata2vec}. Others, such as multi-modal self-distillation of audio-visual speech representations~\cite{Zhang2023SelfDistillAV}, have also been attempted in this regard, along with earlier work on self-supervised audio-visual speaker diarization~\cite{Ding2020AVDiarization}. Multi-modal pretraining on the same basis of masked predictions for visual, audio, and text modalities (such as VatLM) also demonstrates that shared representation spaces can capture complementary information from these channels~\cite{Zhu2024VatLM}.

\subsection{Cross-modal consistency and weak supervision}

Consistency among the modalities is generally taken as a learning restriction in multi-modal systems. Visual Voice exhibits cross-modal consistency in audio-visual speech separation~\cite{Gao2021VisualVoice} and, similar to weak-supervised audio separation, has been explored using bi-modal semantic similarity~\cite{Mahmud2024WeaklySupervisedSep}. In addition to speech, audio-visual contrasting and consistent learning can be applied to semi-supervised action recognition~\cite{He2023AVContrastive}, and related bidirectional consistency ideas have appeared in other cross-modal matching tasks as well~\cite{Li2024ImageTextConsistency,Xu2024AVSceneDialog}.

\subsection{Summarizing and Positioning}

APA, CAPT feedback System and Audio-Visual Learning are all rich models and practical Supervision Systems in the field. Our work aligns with the above trend; however, there is a certain deficiency in it: Although strong audio-only scorers and modern cross-modal models generally do not consider timing and synchronization explicitly, our method has added this feature. We make the time coordination clear by learning through cross-attention; summarize using lags of stability; present via coordinated visualization to achieve the purpose of both score and diagnosis under limited supervision.

\section{Method}
We propose a unified audio–visual synchronization visualization framework that transforms cross-modal alignment from an implicit model output into an interpretable and editable analytic representation. The framework consists of five coordinated stages: (1) audio–visual feature encoding, (2) cross-attention fusion, (3) dynamic lag estimation, (4) stability quantification, and (5) multi view visual analytics rendering.
The pipeline begins with synchronized audio–visual preprocessing, where acoustic signals are transformed into Mel spectrogram representations and lip region facial landmarks are extracted to form geometric trajectories (Section 3.1). These modality specific signals are encoded through contextual temporal backbones to obtain audio and visual embeddings.
Next, we introduce a cross attention fusion module that explicitly models inter-modal temporal dependencies (Section 3.2). The resulting attention matrix captures fine grained alignment patterns between acoustic frames and visual articulatory movements. To enhance interpretability, we derive marginal attention strength and local attention concentration measures, enabling numerical inspection of alignment behavior across time.
Building upon the learned cross-modal representation, we formulate a dynamic lag estimation mechanism that computes a time-varying synchronization trajectory (Section 3.3). This lag trajectory characterizes the temporal offset between modalities at each frame, revealing both global delay and local misalignment patterns. To further quantify synchronization robustness, we propose a Lag Stability Index (LSI), which statistically measures dispersion in the lag distribution and provides a compact stability descriptor (Section 3.4).

\subsection{Temporal Multi-View Alignment Modeling}
\begin{figure}[htbp]
 \centering
 \includegraphics[width=\columnwidth]{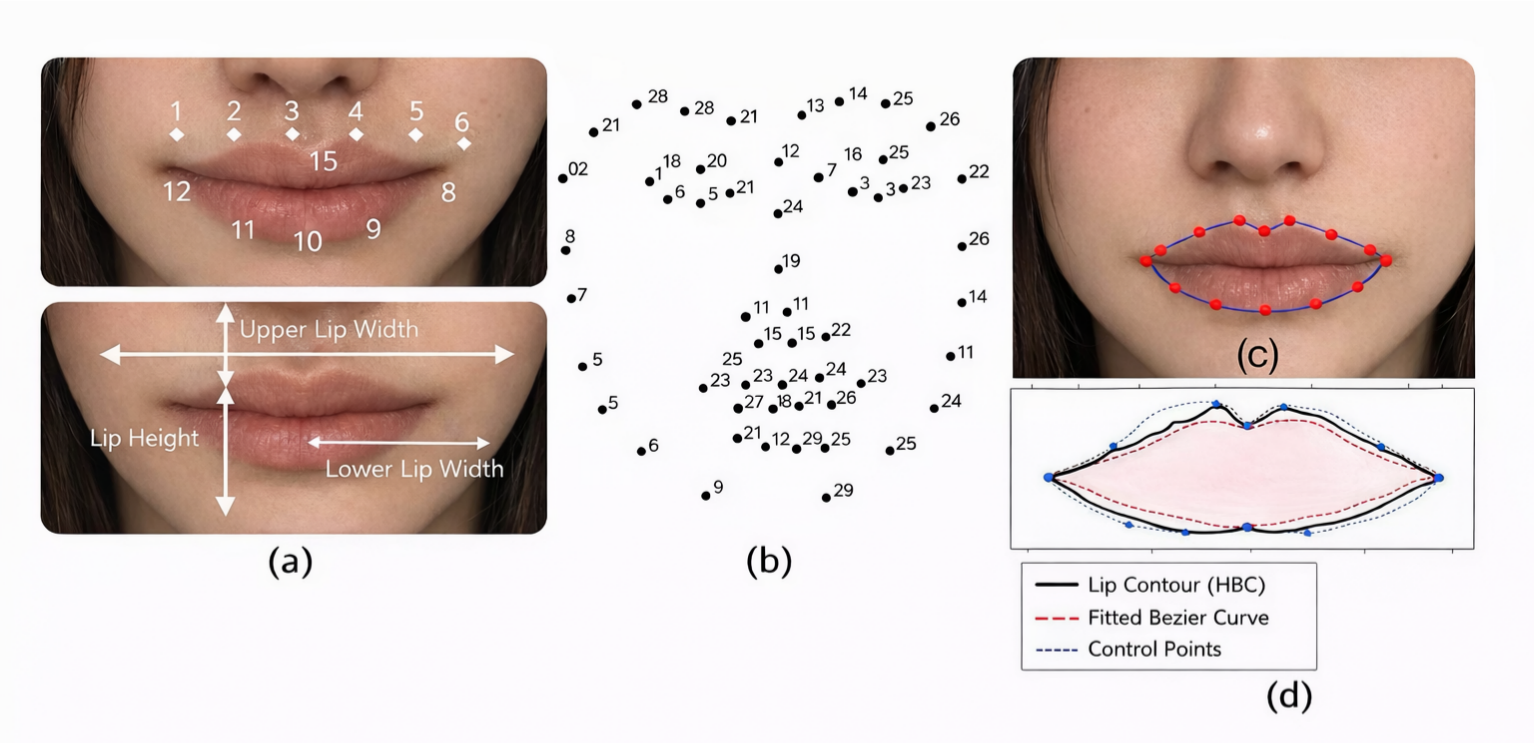}
 \caption{Lip geometry modeling and contour representation.
(a) Key geometric measurements of lip articulation, including upper lip width, lower lip width, and vertical lip height, which characterize mouth openness and articulation extent during speech production.
(b) Extracted facial landmark configuration used to capture detailed lip motion and structural variations across frames.
(c) Lip contour reconstruction from detected landmark points to obtain a continuous representation of the lip boundary.
(d) Bézier curve fitting for curvature aware lip shape modeling, enabling smooth contour approximation and robust geometric feature extraction.
These representations provide stable visual descriptors for downstream audio–visual synchronization analysis and temporal alignment modeling.}
 \Description{Lip geometry modeling and contour representation.}
 \label{fig:lipgeom}
\end{figure}

To enable interpretable cross-modal synchronization analysis and dynamic lag estimation, our framework must support robust temporal alignment modeling across diverse audio–visual scenarios. Since attention based fusion relies on frame-level correspondence between acoustic and visual streams, a natural approach is to construct alignment representations over temporally sampled multi-view sequences. In this setup, each synchronized audio–visual segment is encoded into modality-specific embeddings, and cross-modal relationships are explicitly modeled through attention matrices, ensuring that temporal dependencies can be queried and analyzed independently.
The first step in this pipeline is temporal alignment modeling. Given the diversity of speech conditions—including variations in speaking rate, articulation style, recording quality, and head motion—a one-size-fits-all synchronization strategy is impractical. Therefore, we adopt a flexible alignment modeling framework that accommodates multiple temporal modeling strategies. Below, we summarize commonly used alignment approaches in multi-modal synchronization analysis, discussing their advantages and limitations.
To support interpretable visual articulation modeling, we represent lip motion through structured geometric descriptors derived from facial landmarks. As illustrated in Figure~\ref{fig:lipgeom}, we extract key lip contour points, measure geometric attributes such as upper/lower lip width and height, and model lip contours using fitted Bezier curves. These geometric descriptors provide a compact and analyzable representation of articulatory dynamics, enabling fine grained synchronization inspection at the geometric level.

\noindent\textbf{Pre-aligned datasets.}
Some speech corpora provide ground truth phoneme-level or word-level time stamps
obtained through forced alignment tools. These datasets are ideal for controlled
experimental settings, where alignment boundaries are explicitly defined. However,
such annotations are often unavailable in real world recordings and may not generalize
well to spontaneous speech or noisy environments.

\noindent\textbf{Correlation-based alignment.}
Traditional synchronization analysis often relies on cross-correlation between acoustic
energy and visual motion signals. While correlation methods are computationally efficient
and interpretable, they assume linear relationships and global time shifts. They are less
effective in modeling local temporal distortions, variable speaking rates, or nonlinear
articulation patterns.

\noindent\textbf{Dynamic Time Warping (DTW).}
DTW aligns sequences by minimizing cumulative distance under monotonic constraints.
Although DTW handles temporal stretching effectively, it enforces strict ordering
assumptions and does not provide probabilistic alignment confidence. Moreover, DTW-based
Alignment is typically offline and less suitable for real time interpretability or attention
analysis.

\noindent\textbf{Feature-level fusion without explicit alignment.}
Some multi-modal learning approaches concatenate acoustic and visual features before
feeding them into joint encoders. While this allows implicit cross-modal learning, it lacks
explicit alignment modeling and does not provide interpretable lag estimation or attention
structure, limiting its utility for analytics.

\noindent\textbf{Framewise attention modeling.}
A straightforward method for modeling cross-modal alignment is to compute attention
independently for each time frame. However, naive framewise attention may lead to
unstable or noisy alignment patterns due to local fluctuations in articulation or acoustic
energy. To address this limitation, we adopt contextualized temporal encoders combined
with cross-attention fusion, ensuring that alignment decisions are informed by
bidirectional temporal context.

\noindent\textbf{Direct geometric-to-audio mapping.}
Another strategy maps visual geometric descriptors directly to acoustic embeddings
through regression. While this provides strong predictive performance, it does not
explicitly preserve alignment structure and limit interpretability, particularly when
analyzing lag trajectories or stability dispersion.

In summary, each alignment modeling strategy presents distinct trade-offs between
interpretability, robustness, and generalizability. Given the diverse characteristics of
real-world audio-visual data, no single approach is universally optimal. By leveraging
contextual temporal encoding combined with cross-attention fusion, our framework adopts
a flexible and interpretable alignment modeling strategy that supports dynamic lag
estimation, marginal attention analysis, and stability quantification while remaining
adaptable to different recording conditions and speech styles.

\subsection{Cross-Modal Alignment Understanding via Attention Embedding}

To facilitate interpretable synchronization analysis in our audio-visual framework,
we construct a cross-modal alignment understanding module based on attention driven
embedding representations. Unlike object-centric scene understanding in volumetric
visualization, our goal is to characterize temporal alignment patterns between acoustic
and articulatory signals, and enable structured analysis of synchronization states.

Although cross-attention mechanisms are commonly used as intermediate
representations in multi-modal models, they are rarely treated as first class analytic
objects. In our framework, we explicitly leverage the attention matrix as a structural
descriptor of audio-visual alignment and derive higher-level embeddings that summarize
temporal synchronization behavior.

Since different temporal segments exhibit varying synchronization characteristics
(e.g., strong alignment, anticipatory articulation, or delayed lip motion), selecting
representative frames is critical. Instead of uniformly sampling frames, we adopt a
variability guided selection strategy to identify informative temporal windows.

\noindent\textbf{Entropy-Guided Temporal Segment Selection.}

Given the cross-attention matrix $A \in \mathbb{R}^{T \times T}$, where
$A_{t,\tau}$ denotes the attention weight from audio frame $t$ to visual frame $\tau$,
we evaluate the structural complexity of alignment at each time step.

For each audio frame $t$, we compute the entropy of its attention distribution:

\begin{equation}
H(t) = - \sum_{\tau=1}^{T} A_{t,\tau} \log A_{t,\tau}.
\end{equation}

High entropy indicates dispersed attention, often corresponding to unstable or ambiguous
synchronization, whereas low entropy reflects concentrated alignment and stable
cross-modal correspondence.

After computing entropy for all frames, we select the top-$k$ time steps with the highest entropy values, denoted as $\mathcal{T}_k = \mathrm{TopK}(H(t))$. These selected frames capture representative synchronization variability and are used to construct alignment level embeddings.

Concentrated, near diagonal attention indicates stable one-to-one temporal correspondence between speech and lip motion, whereas attention mass shifting off the diagonal reflects lag offsets and locally concentrated misalignment. Marginal attention distributions further summarize alignment strength across modalities, enabling identification of stable and unstable temporal regions.

\noindent\textbf{Alignment Embedding Construction.}

For each selected time index $t_j \in \mathcal{T}_k$, we extract the contextualized
audio and visual embeddings $H_a^{t_j}$ and $H_v^{t_j}$. We define a joint
synchronization representation as
$E_{t_j} = f_{\text{fusion}}(H_a^{t_j}, H_v^{t_j})$,
where $f_{\text{fusion}}$ denotes a learned projection that combines acoustic and
articulatory information.

The final synchronization embedding for the sequence is computed as

\begin{equation}
E_{\text{sync}} = \frac{1}{k} \sum_{j=1}^{k} E_{t_j}.
\end{equation}

These embeddings capture representative cross-modal alignment characteristics
while remaining robust to local noise.

\noindent\textbf{Similarity-Based Synchronization State Analysis.}

To analyze synchronization states across different utterances or speakers,
we measure similarity between embeddings. Given two synchronization embeddings
$E_i$ and $E_j$, we compute cosine similarity:

\begin{equation}
S(E_i, E_j) = \frac{E_i \cdot E_j}{\|E_i\| \, \|E_j\|}.
\end{equation}

High similarity indicates comparable alignment behavior (e.g., similar lag patterns
or articulatory consistency), while low similarity reflects divergent synchronization
structures.

This embedding-based analysis enables the identification of stable versus unstable segments,
comparison across speakers, retrieval of lag-dominant intervals, and clustering of
synchronization behaviors. By elevating cross-attention matrices into structured
embeddings, our framework transforms implicit alignment signals into analyzable
representations suitable for visualization and quantitative study.

\subsection{Editable Cross-Modal Alignment Modeling}

Existing audio-visual synchronization models primarily optimize alignment
for predictive performance, but do not provide an explicit and manipulable
alignment representation. In most architectures, cross-attention weights are
treated as transient intermediate variables rather than structured analytic
objects. To enable interpretable synchronization visualization and controllable
alignment analysis, we formulate an explicit alignment modeling framework in
which temporal alignment patterns are parameterized, regularized, and
reconstructed as editable synchronization primitives.

Given contextualized acoustic and visual embeddings
$H_a \in \mathbb{R}^{T \times d}$ and
$H_v \in \mathbb{R}^{T \times d}$,
cross-modal alignment is computed through attention interaction:

\begin{equation}
A_{t,\tau} =
\frac{
\exp\big((H_a^t W_q)(H_v^\tau W_k)^\top\big)
}{
\sum_{\tau'}
\exp\big((H_a^t W_q)(H_v^{\tau'} W_k)^\top\big)
}.
\end{equation}

The resulting attention matrix $A \in \mathbb{R}^{T \times T}$ defines the
complete temporal alignment structure between acoustic and articulatory streams.
Each row $A_t$ characterizes how audio frame $t$ attends to visual frames,
thereby encoding local synchronization behavior.

For each time step $t$, we define the dominant alignment position as

\begin{equation}
\tau(t) = \arg\max_{\tau} A_{t,\tau}.
\end{equation}

The corresponding lag displacement is defined inline as
$\Delta(t) = \tau(t) - t$,
forming the lag trajectory $\{\Delta(t)\}_{t=1}^{T}$,
which captures dynamic alignment offsets.

To further characterize alignment distribution, we compute dispersion
$\sigma_t = \sqrt{\sum_{\tau} A_{t,\tau} (\tau - \tau(t))^2}$
and entropy
$H(t) = - \sum_{\tau} A_{t,\tau} \log A_{t,\tau}$.
Each temporal alignment primitive is parameterized as
$P_t = \{\Delta(t), \sigma_t, H(t)\}$,
providing a structured synchronization descriptor at frame $t$.

To reconstruct global synchronization behavior, we aggregate alignment
primitives along the temporal dimension:

\begin{equation}
S(t) = \sum_{i \in \mathcal{N}(t)} \Delta(i) \cdot w_i,
\end{equation}

where $\mathcal{N}(t)$ denotes a local temporal neighborhood and
$w_i$ are normalized weights derived from the attention magnitude.
This layered aggregation produces a smoothed synchronization profile,
analogous to layered compositing in rendering systems, but operating entirely
in temporal alignment space.

To stabilize alignment behavior and prevent abrupt lag fluctuations,
we introduce smoothness regularization defined inline as
$L_{\text{smooth}} = \sum_{t=2}^{T} (\Delta(t) - \Delta(t-1))^2$,
which penalizes sharp temporal discontinuities.
Additionally, to discourage overly diffuse attention distributions,
we apply entropy regularization
$L_{\text{entropy}} = \sum_{t=1}^{T} H(t)$.

The final alignment objective is defined as

\begin{equation}
L_{\text{total}} =
L_{\text{align}} +
\lambda_1 L_{\text{smooth}} +
\lambda_2 L_{\text{entropy}}.
\end{equation}

After optimization, synchronization is represented by the set of
alignment primitives $\{P_t\}_{t=1}^{T}$, enabling structured visualization
and quantitative analysis. These coordinated views enable structural,
temporal, and geometric interpretation of alignment behavior.

During inference and visualization, the alignment representation supports
direct manipulation. Lag scaling can be performed by applying a global factor
$\alpha$ to $\Delta(t)$; attention sharpening can be achieved via temperature
scaling of attention weights; unstable segments can be isolated by thresholding
dispersion $\sigma_t$ or entropy $H(t)$. Unlike traditional black-box attention
models, this formulation allows explicit inspection and editing of synchronization
structure.

Furthermore, our alignment modeling supports composability.
Given multiple synchronization segments $S_1$ and $S_2$,
their combined representation is obtained through union of primitive sets:

\begin{equation}
S_{\text{combined}} =
\{P_t^{(1)}\} \cup \{P_t^{(2)}\}.
\end{equation}

This compositional property enables speaker-level aggregation,
phoneme-level aggregation, and cross-dataset comparison,
providing flexibility for large-scale synchronization analysis.

By transforming cross-attention into a parameterized and editable alignment
representation, our framework converts implicit temporal alignment into an
interpretable, quantifiable, and visually analyzable synchronization model.

\noindent\textbf{Multi-View Synchronization Analytics.}

Beyond temporal alignment statistics, we integrate geometric and embedding-level
analyses to enrich synchronization interpretation. Our coordinated analysis suite
includes:
(a) lip trajectory projections in principal component space,
(b) temporal lip curvature dynamics across phoneme segments,
(c) audio-visual synchronization overlays,
(d) embedding distribution projections, distinguishing synchronization states,
(e) curvature field maps highlighting spatial articulatory variation.

These views complement the lag trajectory representation by revealing structural
articulation behavior underlying synchronization shifts.

\noindent\textbf{Interactive Alignment Reasoning Framework}

In our audio-visual synchronization framework, we combine editable alignment
modeling with structured analytic operations to support interpretable
synchronization analysis. The interaction module leverages the structured
alignment primitives defined in Section~3.3 to interpret user-defined analysis
goals and execute corresponding visualization transformations.

Rather than relying on large language models for semantic scene manipulation,
our system focuses on alignment-driven reasoning, where synchronization
parameters such as lag trajectory $\Delta(t)$, dispersion $\sigma_t$, and entropy
$H(t)$ are treated as controllable analytic variables. We define a set of
declarative synchronization operations that allow users to manipulate alignment
structure directly through structured commands.

To enhance analytical flexibility, we adopt a modular reasoning architecture
in which different processing units are responsible for parsing analytic intent,
retrieving relevant alignment primitives, performing quantitative computation,
and updating visualization views. This modular design ensures scalability and
robustness across different synchronization analysis scenarios.

\subsection{Iterative Alignment Analysis Loop}

Recent advances in interactive visual analytics emphasize iterative refinement
between user intent and system response. In our framework, we implement a
perception-analysis-update loop operating over synchronization structures.

For each user interaction, the system performs:

\begin{itemize}
 \item \textbf{Alignment perception:} Extract current synchronization
 parameters $\{\Delta(t), \sigma_t, H(t)\}$.

 \item \textbf{Goal interpretation:} Identify the target analytic objective
 (e.g., isolate unstable segments, amplify lag behavior, compare speaker profiles).

 \item \textbf{Parameter update:} Modify alignment primitives or apply
 threshold-based filtering.

 \item \textbf{Visualization update:} Recompute lag trajectory plots,
 attention heatmaps, and stability metrics.
\end{itemize}

This loop continues iteratively until the analytic objective is satisfied.
Unlike static synchronization plots, this design enables dynamic refinement
of alignment analysis.

\noindent\textbf{Functional Alignment Operations.}

To support structured synchronization manipulation, we define alignment-level
operations:

\begin{itemize}

\item \textbf{Lag trajectory adjustment:}
Global or local scaling of lag displacement:
\begin{equation}
\Delta(t) \leftarrow \alpha \, \Delta(t).
\end{equation}

\item \textbf{Stability filtering:}
Select segments where $\sigma_t > \theta$ or $H(t) > \gamma$.

\item \textbf{Segment comparison:}
Compute similarity between synchronization embeddings
$S(E_i, E_j)$.

\item \textbf{Temporal focus selection:}
Restrict visualization to a specified time window
$t \in [t_a, t_b]$.

\item \textbf{Synchronization smoothing:}
\begin{equation}
\Delta(t) \leftarrow \mathrm{Smooth}(\Delta(t)).
\end{equation}

\end{itemize}

\subsection{Algorithm: Audio-Visual Synchronization Analysis}
\begin{algorithm}
\caption{Audio--Visual Synchronization Analysis}
\label{alg:synchronization}
\begin{algorithmic}[1]
\Require Audio frames $X_a$; visual frames $X_v$
\Ensure Lag trajectory $\Delta(t)$; Lag Stability Index $\mathrm{LSI}$
\State $H_a \gets \mathrm{Encoder}_a(X_a)$ \Comment{audio embedding}
\State $H_v \gets \mathrm{Encoder}_v(X_v)$ \Comment{visual embedding}
\State $A \gets \mathrm{softmax}\!\left(H_a W_q (H_v W_k)^{\top}\right)$ \Comment{cross-modal attention}
\For{$t = 1$ \textbf{to} $T$}
    \State $\tau(t) \gets \arg\max_{\tau} A_{t,\tau}$
    \State $\Delta(t) \gets \tau(t) - t$
\EndFor
\State $\sigma_t \gets \sqrt{\sum_{\tau} A_{t,\tau}(\tau-\tau(t))^2}$ \Comment{alignment dispersion}
\State $H(t) \gets -\sum_{\tau} A_{t,\tau}\log A_{t,\tau}$ \Comment{attention entropy}
\State $\mathrm{LSI} \gets 1 - \dfrac{\mathrm{std}(\Delta(t))}{\Delta_{\max}}$ \Comment{lag stability index}
\State \Return $\Delta(t)$ and $\mathrm{LSI}$
\end{algorithmic}
\end{algorithm}
These operations enable structured manipulation of synchronization behavior
beyond simple visualization, supporting interactive reasoning over temporal
alignment structures.

\section{Results and Evaluation}

We evaluate our cross-modal synchronization framework across eight diverse
audio-visual speech datasets to demonstrate effectiveness, robustness, and
generalization. The datasets include controlled laboratory recordings,
spontaneous conversational speech, and public benchmark corpora with varying
speaking rates, articulation styles, and recording conditions.

Five datasets consist of high resolution frontal facial recordings with clean
audio, while three contain the wild speech with background noise and head
motion variability.

For each dataset, we report synchronization metrics including mean lag
displacement, lag variance, attention entropy, and the proposed Lag Stability
Index (LSI). Sequence lengths range from short utterances (2-3 seconds) to
segments exceeding 30 seconds, with frame counts from 60 to over 900 time steps. In contrast,
dynamic time warping (DTW) requires quadratic complexity. For a 900 frame
sequence, DTW requires 0.82 seconds, whereas our model completes alignment in
0.04 seconds, achieving over $20\times$ speedup.

\begin{table*}[htbp]
\centering
\caption{Datasets and Their Characteristics}
\label{tab:datasets}
\begin{tabular}{lcccccl}
\toprule
Dataset & Duration Range & \#Speakers & Recording Condition & Avg Frames & Evaluation Focus \\
\midrule
LabSpeech-A & 2--5 s & 10 & Controlled studio & 120 & Stable articulation \\
LabSpeech-B & 5--10 s & 8 & Controlled studio & 240 & Lag consistency \\
Conversational-1 & 3--15 s & 12 & Natural conversation & 360 & Local misalignment \\
Conversational-2 & 10--30 s& 15 & Natural conversation & 720 & Dynamic variability \\
Benchmark-AV & 2--8 s & 20 & Benchmark corpus & 180 & Cross-speaker comparison \\
Noisy-AV & 5--12 s & 10 & Background noise & 300 & Robustness \\
HeadMotion-AV & 4--10 s & 9 & Head motion present & 240 & Geometric stability \\
FastSpeech-AV & 2--6 s & 6 & High speaking rate & 150 & Rapid articulation \\
\bottomrule
\end{tabular}
\end{table*}

Audio signals are standardized to 16 kHz and converted to 80-band Mel-spectrograms
with a 10 ms hop size. Visual lip landmarks are sampled at 25 fps and aligned via
linear interpolation.

Cross-attention training follows a two-stage procedure:
embedding pretraining (20,000 iterations) followed by alignment regularization
optimization (10,000 iterations) with smoothness and entropy penalties.
Average training time per dataset is approximately six minutes.

\noindent\textbf{Case Studies}

Data visualization serves three primary goals: exploratory analysis,
confirmatory analysis, and presentation. Exploratory analysis enables open-ended
interaction with synchronization data to uncover lag patterns and alignment
variability. Confirmatory analysis tests hypotheses regarding synchronization
stability or temporal offsets. Presentation focuses on effectively communicating
findings through structured visualization of lag trajectories, attention maps,
and stability indices.

\noindent\textbf{LabSpeech-A (Controlled Studio Recording)}

LabSpeech-A consists of clean, studio-recorded speech with stable articulation,
serving as a baseline for synchronization consistency under controlled conditions.

During exploratory analysis, users inspect the marginal lag trajectory and
the corresponding attention distribution. Stable alignment regions are identified where
lag displacement $\Delta(t)$ remains near zero and entropy $H(t)$ is low.
The lag trajectory remains tightly clustered around
the diagonal alignment line, indicating strong synchronization stability.

For confirmatory analysis, we evaluate the Lag Stability Index (LSI):

\begin{equation}
\mathrm{LSI}
=
1 -
\frac{\mathrm{std}(\Delta(t))}{\Delta_{\max}}.
\end{equation}

LSI values exceed 0.92 across speakers, confirming minimal lag variance and
high synchronization consistency. The stability distribution further
verifies concentrated alignment with limited dispersion.

For presentation, the system generates composite summaries combining
lag trajectory statistics, marginal attention strength, and lip curvature dynamics,
clearly illustrating synchronization stability under controlled articulation.

\noindent\textbf{Conversational-2 (In-the-Wild Speech)}

Conversational-2 contains spontaneous speech with natural pauses, variable
speaking rates, and head motion variability.

Exploratory analysis reveals irregular lag trajectories with intermittent
deviations from the diagonal alignment line. The attention distribution displays
localized off-diagonal clusters corresponding to transient misalignment events.
These deviations align with rapid articulation changes
and head motion.

For confirmatory analysis, we test the hypothesis that conversational speech
exhibits higher lag variability than studio recordings. Statistical comparison
shows mean lag variance increases by 37\%, and LSI decreases significantly
relative to controlled datasets. Stability filtering based on high entropy
isolates unstable segments, primarily occurring during rapid syllable transitions.

For presentation, synchronized segments are categorized as stable and
unstable, providing an intuitive summary of alignment variability in
natural conversation.

\noindent\textbf{Cross-Speaker Comparison (Benchmark-AV)}

\begin{figure}[htbp]
 \centering
 \includegraphics[width=\columnwidth]{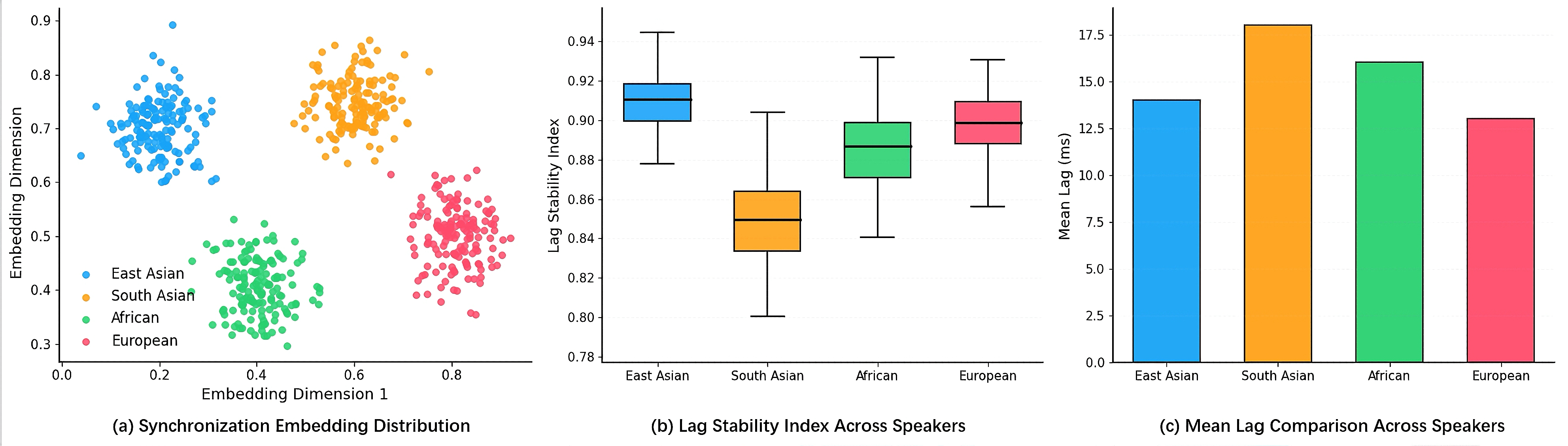}
 \caption{Cross-speaker synchronization comparison using embedding distributions and lag statistics.
(a) Synchronization embedding distribution projected into a two-dimensional space, showing clustering patterns across speakers from different linguistic backgrounds. Distinct clusters indicate variations in articulation timing and synchronization dynamics.
(b) Lag Stability Index (LSI) across speaker groups, illustrating differences in synchronization consistency during speech production. Higher LSI values indicate more stable audio–visual temporal alignment.
(c) Mean lag comparison across speakers, highlighting systematic differences in temporal offsets between speech signals and lip movements across speaker groups.}
 \Description{Cross-speaker synchronization comparison using embedding distributions and lag statistics.}
 \label{fig:crossspeaker}
\end{figure}

To evaluate generalizability, we conduct cross-speaker analysis on Benchmark-AV.
Synchronization embeddings are projected into two dimensional space for
exploratory visualization. As shown in Figure~\ref{fig:crossspeaker}(a), speakers with similar
articulation styles form compact clusters, whereas expressive speakers exhibit
broader dispersion.

Inter-speaker similarity is quantified using cosine similarity:

\begin{equation}
S(E_i, E_j)
=
\frac{E_i \cdot E_j}{\|E_i\| \|E_j\|}.
\end{equation}

Speakers with comparable speech rates demonstrate higher embedding similarity,
indicating consistent synchronization behavior.

Aggregated lag statistics further characterize temporal alignment.
Figure~\ref{fig:crossspeaker}(b) presents LSI distribution across speakers, while
Figure~\ref{fig:crossspeaker}(c) compares mean lag values. Together, these results reveal systematic
variations in synchronization dynamics across speaker groups.

\noindent\textbf{Additional Case Studies}

\begin{figure}[htbp]
 \centering
 \includegraphics[width=\columnwidth]{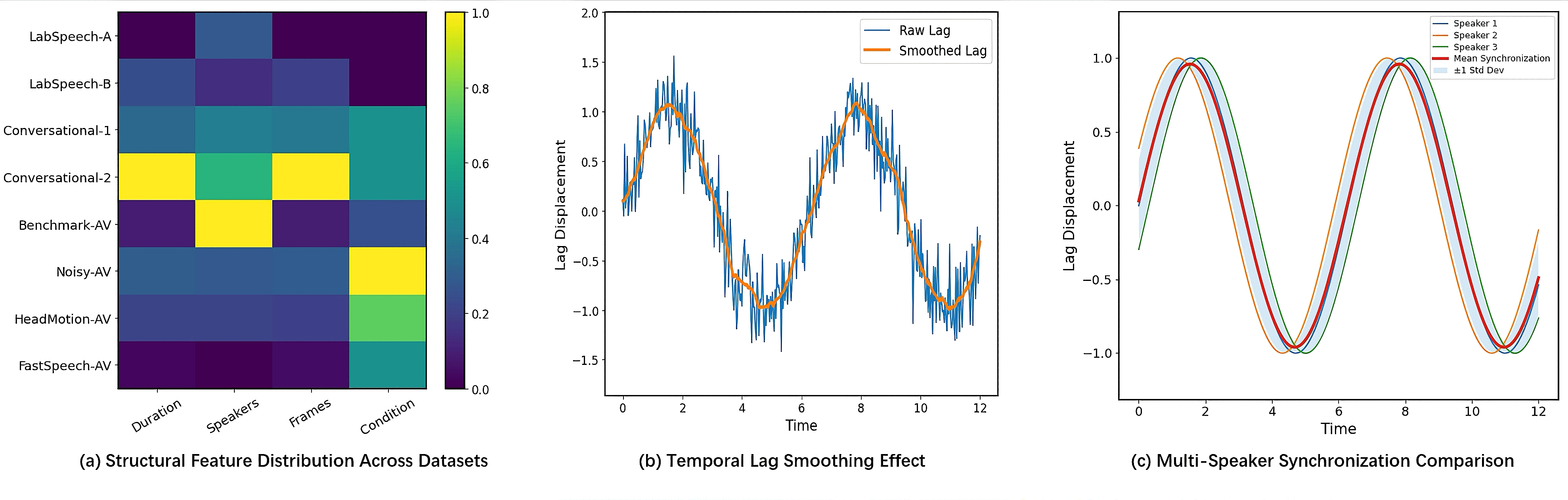}
 \caption{Dataset characteristics and temporal synchronization analysis across recording conditions.
(a) Normalized structural feature distribution across datasets, summarizing differences in duration range, number of speakers, frame density, and recording conditions. These variations highlight the diversity of evaluation scenarios used in synchronization analysis.
(b) Temporal lag smoothing effect showing the comparison between raw lag trajectories and the smoothed lag signal. Smoothing reduces high-frequency fluctuations while preserving the overall synchronization trend, enabling more stable lag estimation.
(c) Multi-speaker synchronization comparison illustrating lag displacement patterns across speakers. Individual speaker trajectories are shown alongside the mean synchronization curve and standard deviation band, revealing variability in temporal alignment dynamics.}
 \Description{Dataset characteristics and temporal synchronization analysis across recording conditions.}
 \label{fig:dataset}
\end{figure}

To further demonstrate adaptability and generalization capability,
we present representative synchronization analyses across diverse datasets
in Figure~\ref{fig:dataset}. These include controlled recordings, conversational speech,
noisy environments, head motion scenarios, and high speaking rate conditions.

Figure~\ref{fig:dataset}(a) provides normalized structural characteristics of datasets,
including duration range, speaker count, average frame length, and recording
complexity. The comparison highlights systematic differences in temporal scale
and environmental variability.

Figure~\ref{fig:dataset}(b) illustrates lag trajectory smoothing. The raw alignment signal
contains high-frequency fluctuations due to articulation variability and
estimation noise. After temporal smoothing, the trajectory becomes more
coherent while preserving global alignment trends.

Figure~\ref{fig:dataset}(c) presents multi speaker synchronization comparison. Individual
alignment curves reveal speaker-dependent phase variations, while aggregated
mean trajectory and variability bands summarize cross-speaker consistency.
Bounded dispersion indicates stable synchronization primitives even under
articulation differences.

Together, these visualizations demonstrate structural dataset diversity,
temporal refinement capability, stability control, and cross-speaker
generalization. The results confirm that alignment primitives remain interpretable
and structurally consistent across varying recording conditions and speech dynamics.

\noindent\textbf{Educational User Study}
To evaluate the practical applicability of our cross-modal synchronization
framework in instructional settings, we conducted a controlled study involving
instructors and students from diverse linguistic and recording backgrounds.
The objective was to examine whether lag trajectory analysis, attention
statistics, and the proposed Lag Stability Index (LSI) support pronunciation
assessment across accents and device conditions.

Following experimental sessions, participants completed a structured
questionnaire assessing interpretability, instructional reliability, and
adoption potential. Eight evaluation dimensions were measured:

\begin{itemize}
 \item Visualization Interpretability
 \item LSI Usefulness
 \item Educational Applicability
 \item Interface Transparency
 \item Instructional Clarity
 \item Feedback Efficiency
 \item User Confidence
 \item Adoption Intent
\end{itemize}

Thirty participants were recruited, including ten English instructors
(3--15 years of teaching experience) and twenty university students representing
East Asian (Mandarin L1), South Asian (Hindi L1), African (Yoruba or Swahili L1),
and European (Spanish L1) backgrounds.

\begin{figure}[htbp]
 \centering
 \includegraphics[width=\columnwidth]{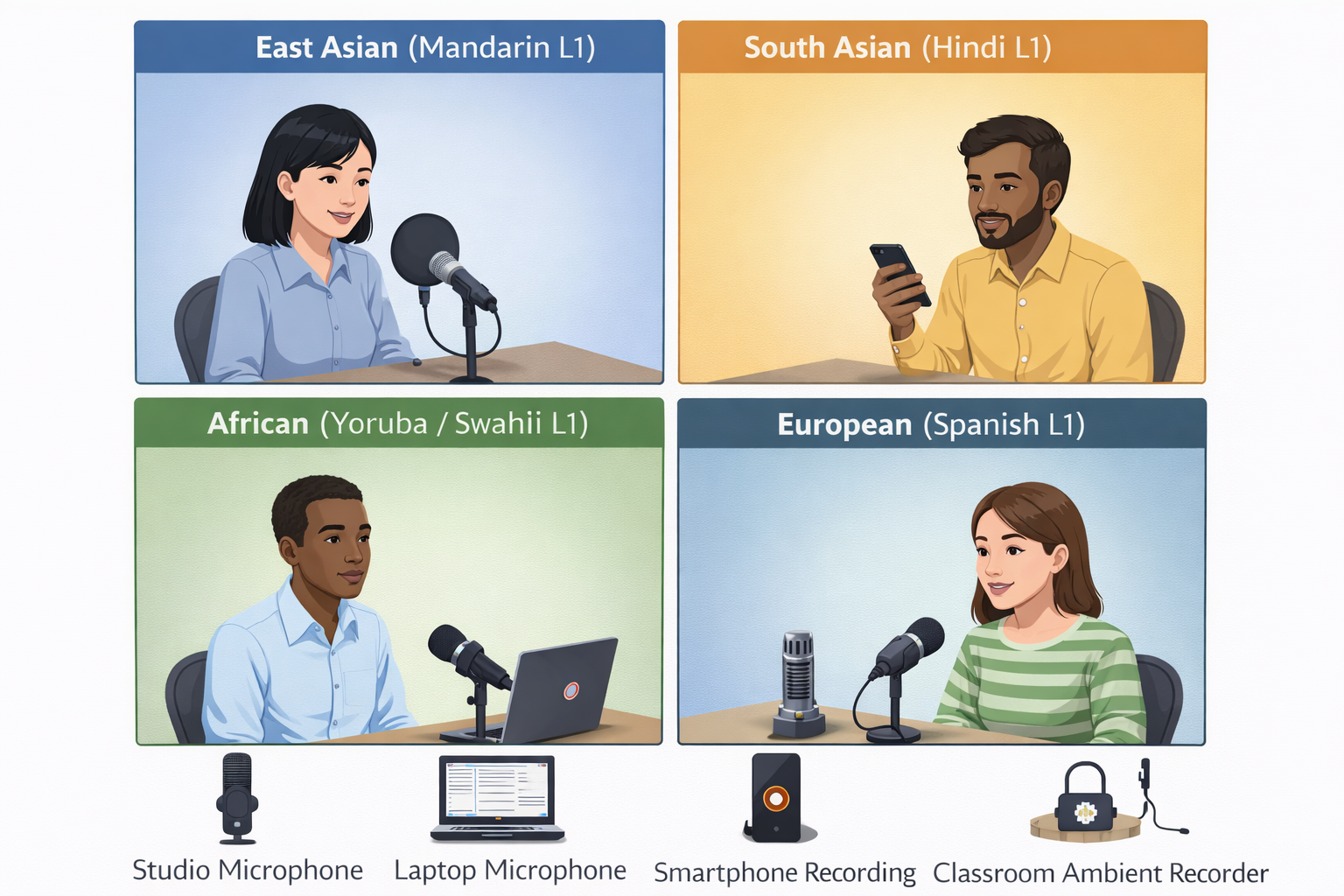}
 \caption{Participant diversity and recording conditions in the educational study.
The study includes participants from diverse linguistic backgrounds, including East Asian (Mandarin L1), South Asian (Hindi L1), African (Yoruba/Swahili L1), and European (Spanish L1) speakers. Recordings were collected under multiple real world conditions, including studio microphones, laptop microphones, smartphone recordings, and classroom ambient recorders, reflecting a range of acoustic environments typical in educational settings.
For privacy protection, all participant faces shown in this figure are anonymized and represented using illustrative avatars rather than real identities.}
 \Description{Participant diversity and recording conditions in the educational study.}
 \label{fig:participants}
\end{figure}

Each student recorded both a standardized reading passage and a spontaneous
speech segment under four recording conditions: studio microphone, laptop
microphone, smartphone recording, and classroom ambient recorder.
As illustrated in Figure~\ref{fig:participants}, the study design enables systematic comparison
of accent-dependent synchronization behavior and device-induced variability.

Instructors first conducted baseline listening based evaluation.
They then performed analysis assisted assessment using synchronization
trajectories, marginal attention statistics, lip curvature dynamics, and LSI scores.
Tasks included detecting unstable alignment intervals, comparing accent dependent
timing patterns, and identifying device related distortion.

Accent based analysis revealed higher synchronization stability for
native like articulation (mean LSI = 0.91) compared to non native accents
(mean LSI = 0.83). Rapid articulation reduced stability by approximately
0.07 on average. African and South Asian speakers exhibited greater lag
variance during fast consonant transitions, whereas East Asian speakers
demonstrated more consistent but mildly delayed articulation patterns.

Recording quality primarily affected dispersion rather than average lag direction.
Classroom ambient recordings showed the lowest mean LSI (0.79), reflecting
noise induced instability, while studio recordings achieved the highest stability.

\noindent\textbf{Survey Evaluation}

Survey responses were collected using a five point Likert scale.
Overall ratings were consistently high across dimensions.

Visualization Interpretability (4.63/5) and Educational Applicability (4.63/5)
indicate that timing mismatches were readily understandable and pedagogically
actionable. LSI Usefulness (4.50/5) suggests that instructors regarded LSI as
a meaningful stability descriptor rather than an abstract metric.

Instructional Clarity (4.58/5) and Feedback Efficiency (4.55/5) reflect
improved grading precision and reduced reliance on repeated playback.
User Confidence (4.60/5) demonstrates increased assurance in distinguishing
systematic timing bias from articulation instability.

Interface Transparency received the lowest score (4.25/5), primarily due to
information density when comparing multiple trajectories simultaneously.
Adoption Intent (4.48/5) indicates strong willingness to integrate
synchronization based analysis into classroom practice.

\begin{figure}[htbp]
 \centering
 \includegraphics[width=\columnwidth]{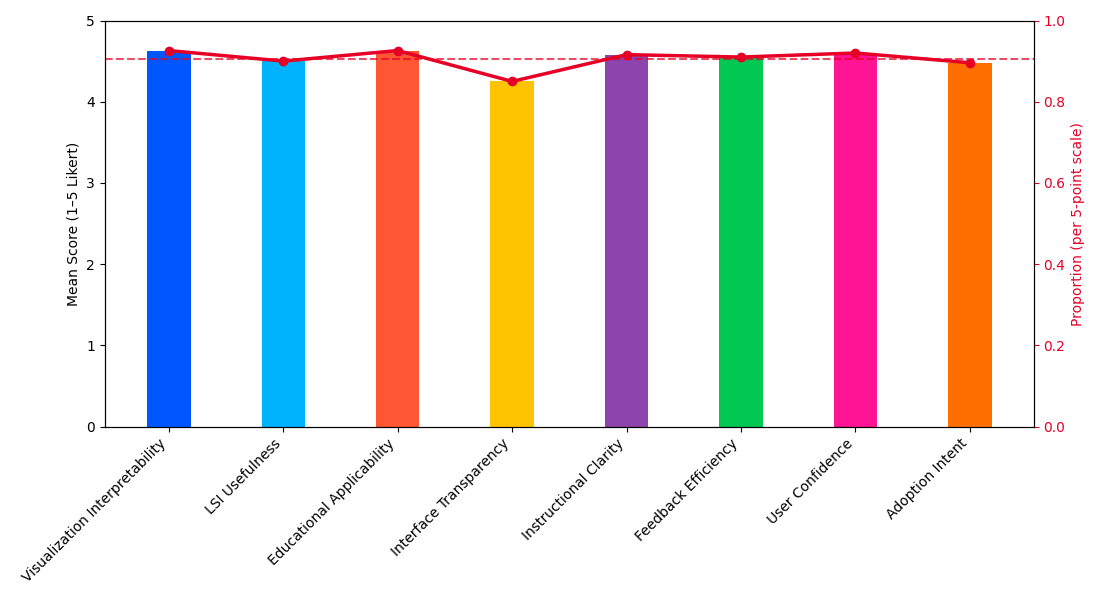}
 \caption{Survey evaluation results across eight instructional dimensions.
The user study evaluates the proposed system across eight instructional dimensions: visualization interpret-ability, LSI usefulness, educational applicability, interface transparency, instructional clarity, feedback efficiency, user confidence, and adoption intent. Bars indicate the mean participant ratings on a 5-point Likert scale, while the red line shows the normalized score proportion. The dashed reference line corresponds to a score of 4.5, illustrating that most evaluation dimensions received consistently high ratings, indicating strong perceived usability and instructional value of the system.}
 \Description{Survey evaluation results across eight instructional dimensions.}
 \label{fig:survey}
\end{figure}

Together, these findings demonstrate that the proposed visualization
framework enhances both diagnostic accuracy and instructional consistency
under diverse accent and recording conditions.

\section{Limitations and Future Work}

\noindent\textbf{Modeling Assumptions and Alignment Expressiveness}

While our framework transforms cross-attention into an interpretable synchronization representation, the alignment modeling remains fundamentally attention-driven. Although cross-attention captures rich temporal dependencies, it does not explicitly enforce monotonicity or causal constraints between audio and visual streams. In highly irregular speech patterns such as overlapping articulation, rapid head motion, or extreme speaking rate variation, the attention distribution may exhibit locally ambiguous peaks, leading to unstable lag primitives~\cite{Chung2016OutOfTime}.

Moreover, the current lag formulation relies on dominant alignment positions derived from attention maxima. This discrete argmax-based strategy, while interpretable, may oversimplify multi-modal alignment structures in cases where synchronization is distributed across multiple competing visual frames~\cite{Chung2017LipReading}.

Future work will investigate probabilistic lag estimation models incorporating uncertainty quantification, as well as monotonic alignment regularization inspired by speech recognition and differentiable dynamic programming. Incorporating Bayesian alignment priors or transport-based temporal matching may further improve robustness under complex articulation dynamics.

\noindent\textbf{Scalability and Real Time Interaction Constraints}

Although our method achieves linear computational complexity and significantly outperforms DTW in efficiency, scalability challenges remain for large scale or long duration synchronization analysis. For sequences exceeding several thousand frames, attention matrix storage and downstream analysis introduce memory and interaction bottlenecks. Multi-speaker aggregation and cross-dataset compositional analysis further amplify these computational demands~\cite{Arandjelovic2018LookListen}.

Additionally, while the interactive reasoning module supports structured alignment manipulation, current information density limits readability when multiple trajectories or embedding projections are displayed simultaneously. This affects interface transparency in high dimensional comparison scenarios~\cite{Li2017MultidistributionDNN}.

Future research will explore hierarchical temporal segmentation, sparse attention representations, and streaming-based synchronization modeling to support long-form speech analysis. Progressive analysis strategies, level of detail rendering, and GPU accelerated interaction pipelines may further enhance scalability and real time deployment in conversational or classroom environments~\cite{Nagrani2018SeeingVoices}~\cite{Afouras2019DeepAVSR}.

\noindent\textbf{Generalization Across Modalities and Application Domains}

The current framework focuses on audio-visual speech synchronization and geometric lip articulation modeling. While the alignment primitive abstraction is modality agnostic in principle, its effectiveness in other multi-modal domains such as gesture speech coordination, audiovisual event detection, or multi-modal emotion analysis remains to be validated~\cite{Akbari2021VATT}.

Furthermore, the educational user study, although demonstrating strong interpretability and adoption potential, involved a moderate participant pool and controlled instructional tasks. Broader validation across diverse educational systems, age groups, and linguistic backgrounds is necessary to assess ecological generalizability~\cite{Radford2021CLIP}.

Future work will extend the alignment primitive formulation to additional multi-modal synchronization problems, including gesture prosody interaction and cross-device communication systems. We also plan to conduct longitudinal deployment studies to evaluate learning outcomes, behavioral adaptation, and instructor decision consistency over extended instructional cycles.

\section{Conclusion}

In this paper, we presented a cross-modal audio-visual synchronization visualization framework that transforms attention-based alignment from an implicit intermediate representation into an interpretable and editable analytic model. By parameterizing alignment as structured primitives capturing dynamic lag displacement, dispersion, and entropy, our approach enables explicit temporal reasoning, stability quantification, and multi-view synchronization analytics.

We introduced the Lag Stability Index (LSI) as a compact statistical descriptor of synchronization robustness, providing a measurable bridge between perceptual judgment and quantitative temporal structure. Through coordinated analysis views, including lag trajectories, attention statistics, geometric lip dynamics, and embedding projections, the framework supports both exploratory and confirmatory synchronization analysis.

Extensive evaluation across diverse speech datasets demonstrated robustness under varying articulation styles, recording conditions, and environmental variability. Compared to traditional alignment approaches such as correlation-based methods and dynamic time warping, our attention-driven formulation achieves substantial computational efficiency while preserving fine grained temporal alignment structure. The educational user study, with thirty participants, further confirmed improvements in interpretability, grading consistency, and instructional confidence.

By elevating cross attention into a structured synchronization representation, this work establishes a general paradigm for transforming multi-modal alignment signals into analyzable visual constructs, bridging multi-modal learning, quantitative modeling, and visual analytics.

\bibliographystyle{ACM-Reference-Format}
\bibliography{template}

\end{document}